\documentclass[10pt,twocolumn,letterpaper]{article}

\usepackage{isba}
\usepackage{times}
\usepackage{epsfig}
\usepackage{graphicx}
\usepackage{amsmath}
\usepackage{amssymb}
\usepackage{mathtools}
\usepackage{tabularx}
\usepackage{colortbl}
\usepackage{hhline}
\usepackage{array}
\usepackage{url}
\usepackage{algorithm}
\usepackage{algcompatible}
\usepackage{algpseudocode}
\usepackage{subfig}
\usepackage{stfloats}
\usepackage{wrapfig}
\usepackage{amsthm}
\usepackage{multirow}

\isbafinalcopy 

\ifisbafinal\fi
\begin{document}

\title{PVSNet: Palm Vein Authentication Siamese Network Trained using Triplet Loss and Adaptive Hard Mining by Learning Enforced Domain Specific Features}

\author{Daksh Thapar, Gaurav Jaswal, Aditya Nigam\\
Indian Institute of Technology Mandi\\
Mandi, India\\
{\tt\small d18033@students.iitmandi.ac.in, gaurav\_jaswal@projects.iitmandi.ac.in, aditya@iitmandi.ac.in}
\and
Vivek Kanhangad\\
Indian Institute of Technology Indore\\
Indore, India\\
{\tt\small kvivek@iiti.ac.in}
}

\maketitle
\thispagestyle{empty}

\begin{figure}[b]
\parbox{\hsize}{\em
2018 IEEE $4^{th}$ International Conference on Identity, Security, and Behavior Analysis (ISBA) \\
978-1-5386-2248-3/18/\$31.00 \ \copyright 2018 IEEE
}\end{figure}

\begin{abstract}
Designing an end-to-end deep learning network to match the biometric features with limited training samples is an extremely challenging task. To address this problem, we propose a new way to design an end-to-end deep CNN framework i.e., PVSNet that works in two major steps: first, an encoder-decoder network is used to learn generative domain-specific features followed by a Siamese network in which convolutional layers are pre-trained in an unsupervised fashion as an autoencoder. The proposed model is trained via triplet loss function that is adjusted for learning feature embeddings in a way that minimizes the distance between embedding-pairs from the same subject and maximizes the distance with those from different subjects, with a margin. In particular, a triplet Siamese matching network using an adaptive margin based hard negative mining has been suggested. The hyper-parameters associated with the training strategy, like the adaptive margin, have been tuned to make the learning more effective on biometric datasets. In extensive experimentation, the proposed network outperforms most of the existing deep learning solutions on three type of typical vein datasets which clearly demonstrates the effectiveness of our proposed method.
\end{abstract}
\begin{figure*}[!htp]
\small
\centering
\includegraphics[width=0.96\linewidth,height=0.23\linewidth]{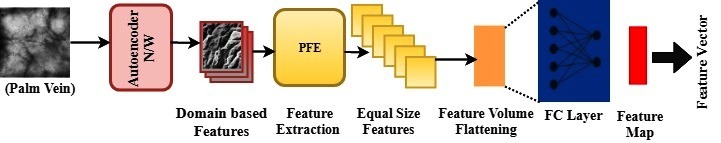} %
\caption{Proposed Palm Vein Feature Extraction (FE) N/W}
\label{fig:aue1a1}
\end{figure*}

\begin{figure*}[!htbp]
\small
\centering
\subfloat{\includegraphics[scale = 0.50]{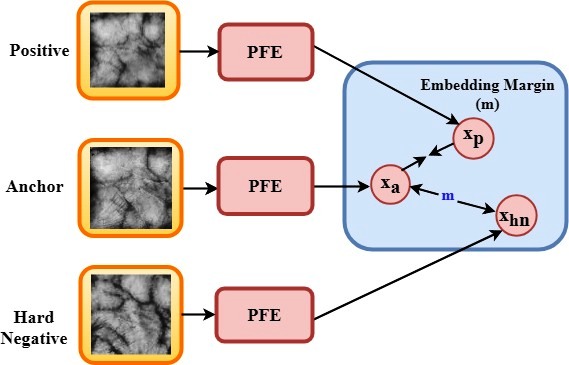}} %
\hspace{1 em}
\subfloat{\includegraphics[scale = 0.50]{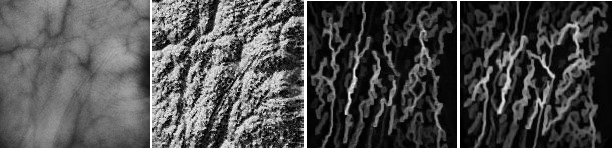}} 
\caption{(a) Proposed Palm Vein Matching N/W; (b) CNN based Image Transformation}
\label{fig:aue1x}
\end{figure*}
\section{Introduction}
With ongoing developments in contact-less imaging sensors, the requirements for vein recognition have been sharply increasing. Besides being unique, the subcutaneous vein structures have the added advantage of lying underneath the skin surface. This makes visibility to the eyes or general purpose cameras difficult and hence, this limits the ease of spoofing if not averting it completely. Back in 2001 \cite{2}, authors found uniqueness in blood vessel networks (under the skin) of the same person and dissimilarities among different individuals and concluded that vein pattern image can be acquired without direct contact with the hand. 
Also, it has been noted that much attention has been paid to the palm-vein and finger-vein modalities individually, but very few researchers attempted to address the problem of presentation attack for vascular technology \cite{44}. However, much recent work has been focused on employing deep convolutional neural networks in various domains, and the field of biometrics is not an exception. 
In a very recent study \cite{47}, authors proposed a two-channel CNN network that has only three convolution layers for finger vein verification. 

\subsection{Challenges and Contribution}
The vein patterns either collected from dorsal or palmer side of the hand show good connected network and provide very vast textural information. The acquisition procedures are convenient and hygienic than other biometric traits. The major open issues in vein verification are the lack of robustness against outdoor illumination, image quality degradation, inconsistent ROI segmentation, and matching between obscure image regions. A few existing image transformation techniques (such as LBP \cite{50}, TCM \cite{49}, IRT\cite{48} etc.) are well proposed in the literature that creates a useful representation of image data and helps to improve the matching task \cite{2}. But no work has been proposed yet that encodes as well as match the image feature through deep learning models. Therefore, efforts have to be made to bridge the gap between deep learning and biometric recognition methods. To best of our knowledge, this is the first attempt in which a convolutional encoder-decoder network has been trained to learn the Texture Code Matrix (TCM) and Image Ray Transform (IRT) based encoding schemes to obtain the domain-specific deep features for palm vein.  

\textbf{Major Contributions :} The major contribution of this work has three folds. (i) An end-to-end deep-learning based vein recognition framework has been designed, which consists of a convolutional encoder-decoder network (CED) and a Siamese network. (ii) We have trained a convolutional deep encoder-decoder network with merged connections for learning the TCM transformation and then trained another similar encoder-decoder network for learning IRT transformation. In this way, we combine both learned transformation models and train an end-to-end CED model from the original image to final IRT transformed image. (iii) Next, a Siamese network has been trained using triplet loss along with hard-negative mining. This Siamese Network is then tested over transformed images obtained from the CED. Thus, the proposed deep learning based vein recognition framework is highly generalized for operating on either of palm vein (CASIA or PolyU or IITI) databases as shown in Figure \ref{fig:aue1aa} and Figure \ref{fig:aue1x}.

\section{Proposed Palm Vein Authentication Siamese N/W}

The proposed biometric authentication approach is based on Siamese convolution neural network framework with a triplet loss function, which enables an idea to learn the distance metric between positive, anchor, and negative embeddings. To do this, first, we enforce an encoder-decoder network to learn domain-specific features, followed by generation of palm vein feature set by stacking inception layers. The complete architecture of proposed system including feature extraction and matching n/w are shown in Figure\ref{fig:aue1a1} and Figure \ref{fig:aue1x}(a). The major importance of our prosed model lies in the end to end learning of the whole system. We use augmentation to contribute the training, select a smaller batch size to converge faster and involve dropouts to prevent from the over-fitting.

\subsection{Palm Vein ROI Extraction}
In order to extract the palm vein ROI's, the hand image is given as input to the region based convolution neural network (CNN) that use different bounding boxes as ground truth to classify and localize the ROI's \cite{52}. Specifically, an input image is first passed through region proposal network (RPN) that provides various probabilistic bounding boxes over which classification and regression heads are applied. Finally, the R-CNN is trained from scratch rather than considering pre-trained weights in order to make out the trained model as problem-specific as possible. The performance of ROI n/w \cite{52}, for segmenting palm vein images in terms of accuracy, at an overlap IOU threshold of $0.5$ is as follows:

\textemdash CASIA \cite{43}: $98.52\%$ 

\textemdash IITI \cite{57}: $99.63\%$

\begin{figure*}[!htp]
\small
\centering
\includegraphics[width=0.97\linewidth,height=0.26\linewidth]{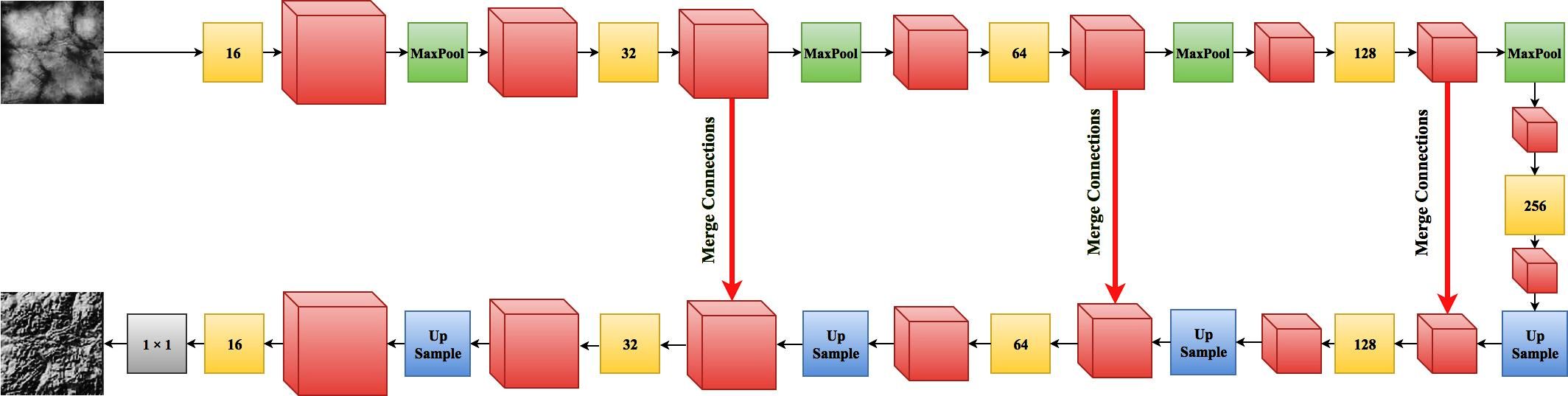} %
\caption{Encoder-Decoder (CED) Architecture. Red blocks depicts $3\times3$ convolutions.}
\label{fig:aue1aa}
\end{figure*}

\subsection{Domain specific transformation learning using Encoder-Decoder Network}
In this work, the encoder-decoder network is inspired by U-Net model which has been widely used for segmentation tasks. We have modified this model for learning image transformations, giving us output similar to what it has been trained on, as shown in Figure \ref{fig:aue1x}(b).

\textbf{Network Architecture :}
The architecture of our deep framework consists of $3\times3$ convolution layers with ReLU activation. Consequently, a $2\times2$ max pooling operation has been used for down-sampling the feature map. Each step in the up-sampling path consists of $2\times2$ up-sampling operation followed by a concatenation with the correspondingly feature map from the contracting path (merge connection). Over the concatenated feature map we apply $3\times3$  convolutions with ReLU activation. The merge connections circumvent the vanishing gradient problem. Figure \ref{fig:aue1aa} shows the network architecture of image transformation model where each red block depicts $3\times3$ convolution layer.
\begin{figure*}
\small
\centering
\includegraphics[width=0.84\linewidth,height=0.2\linewidth]{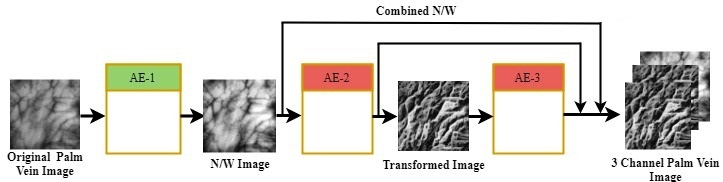} %
\caption{Combined Encoder-Decoder Architecture for Palm Vein Biometrics}
\label{fig:aue2}
\end{figure*}

\textbf{Network Training :}
For training the encoder-decoder networks, we first constructed ground truth by performing a transformation operation namely TCM on 600 palm vein samples and thereafter we applied another transformation IRT over the TCM images. The detailed description of the network training is as follows:

An encoder-decoder network is trained to learn the TCM operation on the original palm vein images. For that, the original image is shown to the network and asked to generate the corresponding TCM image. A second encoder-decoder is trained which takes TCM image as input and has to learn IRT operation on TCM images. Finally, we merge the two networks (end to end)  to create one deep stacked encoder-decoder network whose ultimate task is to render the IRT image from the original image as shown in Figure \ref{fig:aue2}. The combined network is then fine-tuned for generating IRT from original images. The generated TCM and IRT images are concatenated with the original image to form a multi-channel feature which is used for matching. 

\textbf{Usefulness of CNN Learned Features :} The visual feature based appearance for palm vein is depicted in Figure \ref{fig:aue1x}(b), which clearly highlights the curvilinear structures of vein patterns. There are many such algorithms like Gabor filter \cite{56}, CS-LBP \cite{57} already used in vein recognition literature which are robust to illumination and suitable for matching. But these algorithms are usually very slow. Hence to speed up the process, we train an encoder-decoder network which learns such feature transformations. Moreover, once the complete PVSNet is trained in end-to-end fashion, the weights of these networks are updated according to the final objective function. This allows the encoder-decoder networks to generate images that are not exact algorithmic enhancements but images that are even more beneficial for matching than the initial feature transformation. Hence, an encoder-decoder network trained for extracting such features is highly capable of enhancing the most discriminative image features.


\subsection{Proposed Siamese Matching Network architecture: Comparison with FaceNet}
Unlike the previous triplet loss functions like SDH \cite{46} and FaceNet\cite{53}, a triplet Siamese network with an adaptive-margin based hard-negative mining scheme has been introduced. These contribute maximally to train the network and results in faster convergence. In order to match the multi-channel features, we train a Siamese network using a triplet loss function. The network consists of a Feature Extractor (FE) which gives us the $128$ dimensional feature embedding for an image. These feature embeddings are normalized and projected over a zero-centred unit radius hyper-sphere. Over these embeddings, we apply triplet loss to train the FE. Post training, we match the samples using $L_2 $ distance between the feature vectors obtained from the FE. The network used in the arms of the Siamese Network is a stack of inception layers. Most of the spatial information present in the images of Palm is line based and is seen in large chunks. Therefore, the initial layers of the feature-extractor network have large rectangular filters, both in horizontal and vertical orientations. The use of large filters allows us to get very high-level spatial-information from the images, and the rectangularity of these filters captures the most discriminatory features from the images. 

\textbf{Embedding N/W:} The goal of embedding n/w is to map the encoded templates, which may initially have different dimensions, to a joint latent space of common dimensionality in which corresponding images of a given subject have high similarity and dissimilar for all different subjects.

\textbf{Triplet-loss Function:}
We train our network using triplet loss as shown in Figure~\ref{fig:aue1x}(a). The loss-function is designed to counter the adverse effect of the scarcity of data. The triplet loss function ($D(a,p,hn)$), based on the distances between anchor ($a$), positive ($p$) and hard-negative ($hn$) embeddings has two properties - that it uses the hinge-loss function to make sure that a high score for different subjects and low score for the same subjects rewards the network with zero-loss, and that it considers the hard-negative images for the subjects to realize the similarity in images from different subjects. The loss for the anchor-image $a$ and any other image $i$ is given by:
\begin{equation}
	J_{i} = L_{2}^{2}(a,i)
\end{equation}
The triplet-loss function is defined below (where $M$ is margin of hinge loss):
\begin{equation}
  D(a,p,hn) = \frac{1}{2}{max(0,M+J_{p}-J_{hn})}  
\end{equation}

\textbf{Hard-Negative Mining: }\label{hardnegative} 
The network needs to understand that the images of two different subject may be similar to some extent and can cause problems while matching. Therefore, we need to provide hard negative samples to network while training. To compute them for each of the training batches, the network is given random negative pairs and asked to predict the dissimilarity score. Out of these pairs, only those are chosen whose score violates the margin. This process is done in an online fashion before choosing each batch. 

\textbf{Making batches: }
The data is fed to the network in batches of size $90$. Given a subject-pose combination, for each of the batch size samples, we have $3$ images: 1) The anchor image is the one corresponding to the given subject-pose combination. 2) The positive image is that of a different pose from the same subject. 3) The negative image is one of the hard-negatives for the given subject-pose combination. The number of positive samples has been increased by using augmented images made using the Augmentor Tool \cite{42}, to supplement the low number of positive samples for a given subject.

\textbf{Tuning the triplet-loss margin: }
The most important hyper-parameter while training the combined model is the margin by which the triplet-loss tries to separate the negative-pair from the positive one. We dynamically increase this margin from $0.2$ to $0.5$ as the training proceeds. 


\textbf{Feature Extractor (FE):} For extracting the feature embeddings of palm vein, we use feature extractor as shown in Figure \ref{fig:aue1a1} (a). 
The detail about the architectures of palm CNN is given in Table \ref{tab:1}. The CNN gives us a $(7*7*512)$ dimensional feature vector that is flattened. 


\textbf{Network Training:} 
To avoid overfitting, we pretrain the CNN of the siamese network as an autoencoder. The input to encoder being the multi-channel feature image and output is a feature map of size $(7\times 7\times 512)$. The decoder has been designed as a mirror of the CNN encoder, which will take the feature map from the encoder as an input and output the multi-channel feature image. This autoencoder is trained over the gallery dataset to ensure that all the generative features present in the images are learned. Once the autoencoder is trained, we discard the decoder, save the weights of the encoder and use it as a pre-trained weights for the CNN in the feature extractors. The detail about the architectures of palm CNN is given in Table \ref{tab:1}. We add a fully connected layer to get the final $128$ dimensional feature embedding. Later we fine tune the FE for the discrimination task using the triplet loss (as discussed above). For initial training, the layers from the CNN in the FE are kept frozen and only the fully-connected layer are trained. This is done to prevent the heavy-loss from untrained fully-connected layer from distorting the generative feature-maps learn by the CNN. Once the loss for the FC layers gets stabilized, all the layers of the network are trained. Finally, we combine both the encoder-decoder pipeline and the Siamese Network to train this whole network in an end-to-end fashion.

\begin{table*}
\centering
\caption{ROC based Performance Analysis over Vein Image Datasets}
\label{tab:veintable1}
\begin{tabular}{|l|l|l|l|l|l|}
\hline
\textbf{Description}   & \textbf{Genuine Matching}  & \textbf{Impostor Matching}  & \textbf{EER} (\%)  & \textbf{CRR} (\%)  & \textbf{DI}  \\ \hline
\multicolumn{6}{|c|}{Intra Vein Matching (CASIA MS Database): Training and Testing on same Images} \\ \hline
Siamese N/W    &       1800               &       358200             &     3.71      & 85.16          & 2.33    \\ \hline
FaceNet  &     1800     & 358200  & 5.77   & 77.16           &  2.11   \\ \hline
\multicolumn{6}{|c|}{Intra  Vein Matching (IITI Vein Database): Training and Testing on same Images}                \\ \hline
Siamese N/W   &     3330   &    1228770  & 0.93     & 97.47  & 2.66    \\ \hline
FaceNet  &   3330    & 1228770    & 2.20     & 93.69  & 2.35    \\ \hline
\multicolumn{6}{|c|}{Intra Vein Matching (PolyU MS Database ): Training and Testing on PolyU Images}                             \\ \hline
Siamese N/W      & 18000   & 8982000      &     0.66      &98.78    & 2.35    \\ \hline
FaceNet    & 18000  & 8982000    & 1.42    & 96.43  & 2.09    \\ \hline
\multicolumn{6}{|c|}{Inter Vein Matching: Training on IITI and Testing on CASIA Images}                                      \\ \hline
Siamese N/W     & 1800   & 358200    & 9.37    &72.33     & 1.92     \\ \hline
FaceNet    & 1800   & 358200   & 12.72  & 57           &  1.69   \\ \hline
\multicolumn{6}{|c|}{Inter Vein Matching: Training on CASIA and Testing on IITI Images}                                      \\ \hline
Siamese N/W      & 3330      & 1228770  & 5.08    & 92.88   & 2.10    \\ \hline
FaceNet   & 3330   & 1228770  & 6.30   & 89   & 2.03     \\ \hline
\multicolumn{6}{|c|}{Inter Vein Matching: Training on PolyU and Testing on CASIA Images}                                      \\ \hline
Siamese N/W      &    1800               &  358200                  &     11.44      &    86.16       & 1.79    \\ \hline
FaceNet          &  1800               &  358200        &    13.84       &   82.66        &   1.73  \\ \hline
\multicolumn{6}{|c|}{Inter Vein Matching: Training on PolyU and Testing on IITI Images}                                      \\ \hline
Siamese N/W    & 3330   & 1228770    & 7.36  & 89.90          &  1.92   \\ \hline
FaceNet   & 3330    & 1228770      & 9.43      & 85.58    & 1.83     \\ \hline
\end{tabular}
\end{table*}

\begin{table}
\centering
\caption{Siamese CNN Architecture}
\label{tab:1}
\begin{tabular}{|c|c|}
\hline
\multicolumn{2}{|c|}{\textbf{Palm Vein CNN}} \\ \hline
\multicolumn{2}{|c|}{Input Size $150\times150$}             \\ \hline
2D Conv (9,3)         & 2D Conv (3,9)        \\ \hline
\multicolumn{2}{|c|}{Maxpool}                \\ \hline
2D Conv (7,3)         & 2D Conv (3,7)        \\ \hline
\multicolumn{2}{|c|}{Maxpool}                \\ \hline
2D Conv (5,3)         & 2D Conv (3,5)        \\ \hline
\multicolumn{2}{|c|}{Maxpool}                \\ \hline
\multicolumn{2}{|c|}{2D Conv (3,3)}          \\ \hline
\multicolumn{2}{|c|}{Maxpool}                \\ \hline
\multicolumn{2}{|c|}{2D Conv (3,3)}          \\ \hline
\multicolumn{2}{|c|}{2D Conv (3,3)}          \\ \hline
\multicolumn{2}{|c|}{Output Size $7\times7\times512$}            \\ \hline
 \end{tabular}
\end{table}

\begin{table*}
\centering
\caption{Comparative Performance Analysis}
\label{tab:compareT}
\begin{tabular}{|l|l|l|l|}
\hline
\multirow{2}{*}{Algorithms}             & \multirow{2}{*}{Database Used} & \multicolumn{2}{c|}{Palm Vein (\%)} \\ \cline{3-4} 
                                        &                                & EER      & CRR     \\ \hline
\multicolumn{4}{|c|}{Handcrafted Feature based State-of-Art Algorithms}                                                                                                                 \\ \hline
WLD \cite{31}          & CASIA MS                    & 6.08           & 97.50             \\ \hline
MPC \cite{44}         & CASIA MS                     & 1.83           & NA                  \\ \hline
NMRT \cite{34}        & CASIA MS                     & 0.51           & 99.17          \\ \hline
Deep-Matching \cite{57}    & CASIA MS              &  2.61            &   NA \\ \hline
 \cite{56}        & PolyU MS                     & 0.26 (NIR)         & NA          \\ \hline
\multicolumn{4}{|c|}{CNN Feature based State-of-Art Algorithms}                                                                                                                         \\ \hline
SDH \cite{46}          & PolyU Finger Vein              &       9.77         &       NA           \\ \hline
Triplet Loss \cite{46} & PolyU Finger Vein              &     13.16           &        NA        \\ \hline
Deep-Vein \cite{32}    & PolyU Finger Vein              &  3.02              &     NA \\ \hline
FaceNet \cite{53}      & CASIA MS                     &   5.77           &      77.16                 \\ \hline
FaceNet \cite{53}     & PolyU MS                     &    1.42          &       96.43        \\ \hline
   
\multicolumn{4}{|c|}{Proposed CNN based State-of-Art PVSNet}                                                                                                                          \\ \hline
PVSNet     & CASIA MS                     &     3.71         &       85.16        \\ \hline
PVSNet    & IITI Vein                   &     0.93         &       97.47        \\ \hline
PVSNet     & PolyU MS                     &      0.66          &       98.78        \\ \hline
 \end{tabular}
\end{table*}
\section{Experimental Analysis}
In distinguishing experiments, the performance of the proposed method has been evaluated in terms of three performance parameters $viz.$ EER (Equal Error Rate), CRR (Correct Recognition Rate) and DI (Decidability Index). We conduct experiments more in-depth on publicly available databases (CASIA multispectral palm print \cite{43}, IIT Indore Hand Vein (available on request), and PolyU multispectral (NIR) palm print \cite{54}) to validate the proposed framework. For each test, the following two testing protocols are defined. \textbf{(a) Intra Vein Testing:} PVSNet has been trained and tested on same vein dataset. For example, CASIA on CAISA, IITI on IITI, and PolyU on PolyU. \textbf{(b) Inter Vein Testing:} PVSNet has been trained over any one of the three vein datasets while tested on the data from other two datasets. For example, CASIA on IITI, CASIA on PolyU, IITI on CASIA, and IITI on PolyU. In addition, another state-of-art deep network architecture named as FaceNet has been tested using same testing protocols, tweaked and fine-tuned as required. 

\textbf{Database Specifications :} The proposed system has been tested on publicly available CASIA multispectral palm print database \cite{43}, IIT Indore Hand Vein database, and PolyU multispectral (NIR) palmprint database. The left and right-hand samples from a subject are considered belonging to separate individuals, resulted in 200 (CASIA), 370 (IIT Indore) and 500 (PolyU) subjects respectively. For each subject, the first half samples are considered as the gallery and the remaining as the probe. 

\textbf{Experiment-1:} In the first test, the comparative potential of proposed PVSNet, as well as FaceNet, has been evaluated by using intra vein matching scheme. The corresponding ROC characteristics w.r.t CASIA, IITI and PolyU datasets are shown in Figure \ref{fig:veria} (a,b). All the details about genuine/ impostor matchings as well as EER, CRR and DI corresponding to the best performance for each dataset are highlighted in Table \ref{tab:veintable1}. On account of results, the following direct observations can be made. (1) The palm vein samples of PolyU and IITI datasets obtains promising verification results than CASIA datasets, using either of CNN model. Since CASIA database allows a high degree of variation of hand postures than PolyU and IITI samples. (2) In overall, the proposed Siamese N/W outperforms FaceNet in terms of achieving lower EER values for each one of palm-vein datasets. Specifically, the Siamese N/w achieves the best individual performance of 0.66 \% EER, and 0.93\% EER on PolyU and IITI images respectively. Since Siamese networks trained with the feature embeddings produced by the autoencoder achieve better matching performance compared to the triplet-based loss function, but the dependence of the Siamese network on pairwise inputs is a crucial issue during inference in terms of complexity. 

\textbf{Experiment-2:} In this test, testing strategy harder than previously used intra vein matching is devised to make a more fair comparison in between the PVSNet and FaceNet models. Using this testing, we run 8 combinations of training and testing sets, where we use one dataset for training and another for testing. In other words, matching is performed similarly to zero shot learning that involves gallery and probe images from test dataset only. 
In this test, one can clearly observe that as we shift from intra to intervein testing scheme, the performance drops abruptly in all studied cases. The complete detail about performance parameters is given in Table \ref{tab:veintable1} and the respective ROC graphs are depicted in Figure \ref{fig:veria} (a,b). The reason is that the model is trained and tested on entirely different vein datasets and the images may be dissimilar in size or shape. For instance, training on CASIA and testing on IITI or testing on IITI and training on CASIA shows better results than the same tests with PolyU images because PolyU vein images are of size $128 \times 128$, while CASIA and IITI are both of same size ($150 \times 150$). Also, the physical configuration of PolyU vein images is different from CASIA and IITI images. Thus, we have not performed inter vein testing on PolyU images. 
\begin{figure*}[!htbp]
\small
\centering
\subfloat{\includegraphics[scale = 0.5]{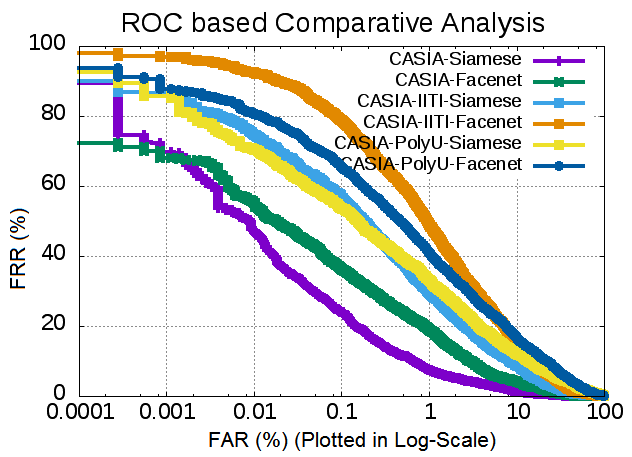}} %
\subfloat{\includegraphics[scale = 0.5]{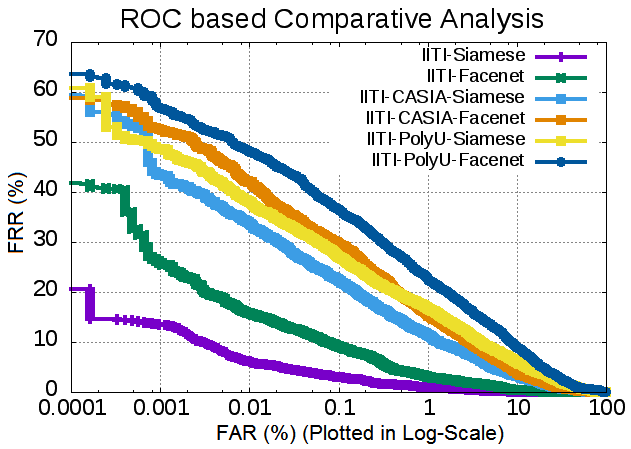}} 
\caption{ROC Performance Analysis: (a) Testing with CASIA; (b) Testing with IITI}
\label{fig:veria}
\end{figure*}

\textbf{Experiment-3:} To illustrate the effectiveness of the proposed PVSNet model, a performance comparison with three handcrafted and four CNN based methods is given in Table \ref{tab:compareT}. The various existing approaches based on deep networks incorporate a CNN based features and trained using the triplet-based loss function which is not perfect of achieving a smaller intra-class variation and a larger inter-class variation with significant performance on the testing set. In contrast to this, the proposed method straightly trains its output to be a compact 128-D embedding using a triplet loss function. A lower value of EER i.e., 0.66\% has been achieved on PolyU palm vein, which is superior to the EER values obtained from any of the CNN based methods \cite {32, 47, 53}. In addition, an EER of 3.71\% is achieved on the most challenging CASIA dataset which is extremely better than the similar works done in. This justifies the strength of network learned feature representation and Siamese based triplet loss network. One can observe that we have achieved state of the art results in almost all the testing strategies adopted in this paper.

\begin{figure*}[!htbp]
\small
\centering
\subfloat{\includegraphics[scale = 0.5]{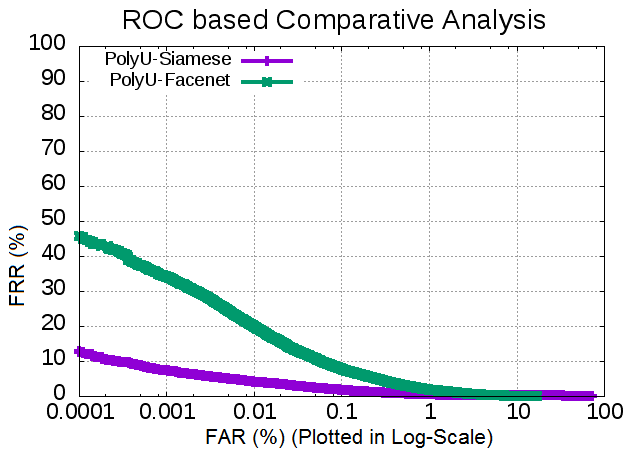}} %
\hspace{1 em}
\subfloat{\includegraphics[scale = 0.5]{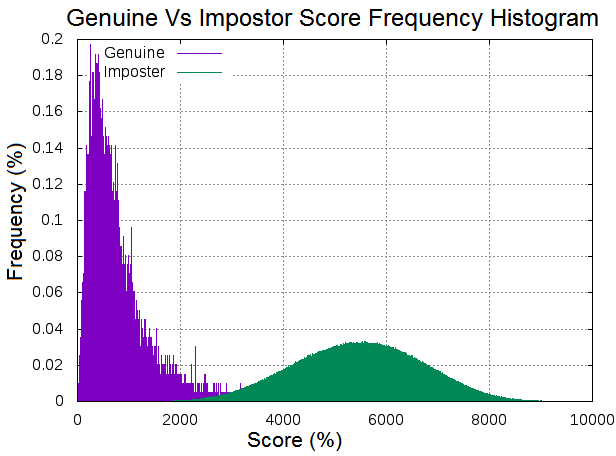}} %
\caption{ROC based Performance Analysis: (a) Testing with PolyU; (b) Genuine Vs Impostor }
\label{fig:veria}
\end{figure*}


\section{Conclusion}
In this work, we have proposed a novel end-to-end deep network design by combining domain-specific knowledge and deep learning representation. The various challenging issues related to vein biometrics have been addressed suitably. The fixed size ROI images have been given to an end-to-end CED augmented with Siamese network trained using triplet loss for vein recognition. 
As a part of future work, we will try to incorporate CNN based ROI segmentation and ROI enhancement network to further improve the recognition performance of the proposed system.
{
\bibliographystyle{ieee}
\bibliography{gaurav}

\begin{thebibliography}{10}\itemsep=-1pt

\bibitem{43}
Casia multi spectral palmprint v1 database.
\newblock \url{http://biometrics.idealtest.org/dbDetailForUser.do?id=6}.

\bibitem{42}
Data augmentor.
\newblock \url{http://augmentor.readthedocs.io/en/master/}.

\bibitem{54}
Polyu multi spectral palmprint database.
\newblock
  \url{http://www4.comp.polyu.edu.hk/~biometrics/MultispectralPalmprint/MSP.htm}.

\bibitem{57}
S.~Bhilare, G.~Jaswal, V.~Kanhangad, and A.~Nigam.
\newblock Single-sensor hand-vein multimodal biometric recognition using
  multiscale deep pyramidal approach.
\newblock {\em Machine Vision and Applications}, pages 1--18, 2018.

\bibitem{44}
J.~H. Choi, W.~Song, T.~Kim, S.-R. Lee, and H.~C. Kim.
\newblock Finger vein extraction using gradient normalization and principal
  curvature.
\newblock In {\em Image Processing: Machine Vision Applications II}, volume
  7251, page 725111. International Society for Optics and Photonics, 2009.

\bibitem{48}
A.~H. Cummings, M.~S. Nixon, and J.~N. Carter.
\newblock The image ray transform for structural feature detection.
\newblock {\em Pattern Recognition Letters}, 32(15):2053--2060, 2011.

\bibitem{47}
Y.~Fang, Q.~Wu, and W.~Kang.
\newblock A novel finger vein verification system based on two-stream
  convolutional network learning.
\newblock {\em Neurocomputing}, 2018.

\bibitem{31}
B.~Huang, Y.~Dai, R.~Li, D.~Tang, and W.~Li.
\newblock Finger-vein authentication based on wide line detector and pattern
  normalization.
\newblock In {\em Pattern Recognition (ICPR), 2010 20th International
  Conference on}, pages 1269--1272. IEEE, 2010.

\bibitem{2}
S.~Im, H.~M. Park, Y.~W. Kim, S.~C. Han, S.~W. Kim, C.~H. Kang, and C.~K.
  Chung.
\newblock An biometric identification system by extracting hand vein patterns.
\newblock {\em Journal-Korean Physical Society}, 38(3):268--272, 2001.

\bibitem{52}
R.~R. Jha, D.~Thapar, S.~M. Patil, and A.~Nigam.
\newblock Ubsegnet: Unified biometric region of interest segmentation network.
\newblock {\em arXiv preprint arXiv:1709.08924}, 2017.

\bibitem{32}
H.~Qin and M.~A. El-Yacoubi.
\newblock Deep representation-based feature extraction and recovering for
  finger-vein verification.
\newblock {\em IEEE Transactions on Information Forensics and Security},
  12(8):1816--1829, 2017.

\bibitem{53}
F.~Schroff, D.~Kalenichenko, and J.~Philbin.
\newblock Facenet: A unified embedding for face recognition and clustering.
\newblock pages 815--823, 2015.

\bibitem{49}
S.~Umer, B.~C. Dhara, and B.~Chanda.
\newblock Texture code matrix-based multi-instance iris recognition.
\newblock {\em Pattern Analysis and Applications}, 19(1):283--295, 2016.

\bibitem{46}
C.~Xie and A.~Kumar.
\newblock Finger vein identification using convolutional neural network and
  supervised discrete hashing.
\newblock In {\em Deep Learning for Biometrics}, pages 109--132. Springer,
  2017.

\bibitem{50}
X.~Yan, W.~Kang, F.~Deng, and Q.~Wu.
\newblock Palm vein recognition based on multi-sampling and feature-level
  fusion.
\newblock {\em Neurocomputing}, 151:798--807, 2015.

\bibitem{56}
D.~Zhang, Z.~Guo, G.~Lu, L.~Zhang, and W.~Zuo.
\newblock An online system of multispectral palmprint verification.
\newblock {\em IEEE transactions on instrumentation and measurement},
  59(2):480--490, 2010.

\bibitem{34}
Y.~Zhou and A.~Kumar.
\newblock Human identification using palm-vein images.
\newblock {\em IEEE transactions on information forensics and security},
  6(4):1259--1274, 2011.

\end{thebibliography}
}

\end{document}